\documentclass{article}
\usepackage[main,final]{neurips_2026}

\usepackage[utf8]{inputenc}
\usepackage[T1]{fontenc}
\usepackage{hyperref}
\usepackage{url}
\usepackage{booktabs}
\usepackage{amsfonts}
\usepackage{amsmath}
\usepackage{amssymb}
\usepackage{nicefrac}
\usepackage{microtype}
\usepackage{xcolor}
\usepackage{graphicx}
\usepackage{placeins}
\usepackage{wrapfig}
\usepackage{tcolorbox}
\usepackage{enumitem}
\usepackage{comment}
\usepackage{tikz}
\usetikzlibrary{positioning,arrows.meta}
\usepackage{pgfplots}
\pgfplotsset{compat=1.18}
\definecolor{rlgreen}{HTML}{386440}
\definecolor{sftgray}{HTML}{858585}
\usepackage{xcolor}
\title{How New Reasoning Strategies Emerge in RL Post-Training: A Controlled Study}

\author{%
  Azwar Abdulsalam\thanks{Corresponding author: \texttt{azwar.azwar.25@ucl.ac.uk}} \quad
  Nishil Patel \quad
  Andrew Saxe \\
  Gatsby Computational Neuroscience Unit, University College London \\
  London, United Kingdom
}

\begin{document}
\maketitle

\begin{abstract}
Does reinforcement learning (RL) post-training with verifiable rewards build
new reasoning procedures, or merely reweight behaviors already latent in the
base model? We study this question in a fully observable rewrite-grammar
environment where the pretraining distribution is known and every generated
rewrite can be audited. A Transformer is pretrained on primitive symbol-rewrite
chains and post-trained with only a binary final-answer reward. RL solves
held-out problems that remain rarely solved by the pretrained model even under
much larger sampling budgets, while rejection fine-tuning (RFT) improves early
but plateaus. Trace analysis reveals a phased mechanism: RL first strengthens
primitive reductions, then enters a chunking phase, composing them into valid
compressed procedures---macro contractions that collapse sequential reductions and
parallel contractions that combine independent ones---which are reused and
consolidated into a stable repertoire. RL's advantage over RFT lies not in
exploration volume but in selectivity: RFT produces many invalid shortcut-like
rewrites, whereas RL concentrates exploration into valid reusable structure.
Pretraining ablations show that strategy emergence depends on whether
pretraining organizes primitive competence into reduction procedures that RL can
later compress. The central result carries over to a pretrained LLM: on Boolean
circuit simplification, GRPO makes valid compressed solutions accessible to
Qwen3-4B at pass@16 that its starting policy rarely produces at pass@1024, and
the learned macros transfer to new variable bindings and to a schema never
required during training. The base model provides weak procedural ingredients;
RL builds them into reliable higher-level strategies.
\end{abstract}

 \section{Introduction}

Does reinforcement learning (RL) post-training teach language models genuinely
new strategies, or does it primarily reweight behaviors already latent in the
base model? Recent work has sharpened this into an active debate. Yue
et al.~\citep{yue2025beyondbase} argue that RL with verifiable rewards often
improves small-$k$ success without expanding the large-$k$ capability frontier;
ProRL~\citep{liu2025prorl} and Yuan et al.~\citep{yuan2025compose} present
evidence that prolonged RL or controlled compositional tasks can uncover
behaviors inaccessible to the base model under extensive sampling; and
imitation-style baselines such as rejection fine-tuning are sometimes
surprisingly competitive~\citep{chu2025sft,xiong2025minimalist}.

A central obstacle is that the relevant mechanisms are hard to observe in
pretrained language models. When a post-trained model exhibits a new behavior,
it is usually unclear whether that behavior was absent from the base model,
merely low-probability, or already present in pretraining data. Aggregate
metrics such as pass@$k$ leave three questions open: what strategies are being
used, whether their emergence reflects broader exploration or selective
filtering, and what pretraining structure makes them reachable in the first
place.

We introduce a fully observable rewrite-grammar environment for studying these questions. The task abstracts a common structure in step-by-step reasoning: useful strategies often compress several local transformations into a reusable procedural chunk. A Transformer is pretrained
from scratch on primitive rewrite chains and then post-trained on a
goal-directed contraction task. Because every generated rewrite can be audited
against the grammar, we can decompose behavior into primitive rule use, valid
procedural chunks---\emph{macro} contractions that compress sequential
reductions and \emph{parallel} contractions that combine independent ones---and
spurious invalid rewrites. This turns normally opaque post-training behavior
into exactly classifiable trajectories, letting us ask whether RL reorganizes
primitive competence into reusable higher-level strategies. Finally, we test the
central result in a pretrained LLM, Qwen3-4B, on Boolean circuit simplification.

\paragraph{Contributions.}
\begin{enumerate}[leftmargin=1.4em,itemsep=2pt,topsep=2pt]
    \item \textbf{A fully observable rewrite-grammar testbed.}
    We introduce an environment in which primitive rules, pretraining histories,
    and generated rewrites are exactly inspectable, enabling a four-way
    taxonomy: primitive, macro, parallel, and spurious.

    \item \textbf{Finite-budget frontier expansion.}
    We show that RL solves held-out problems the pretrained model rarely solves
    even under much larger sampling budgets, with gains that emerge late and
    grow with difficulty.

    \item \textbf{Phased procedural chunking.}
    We show that RL first strengthens primitive reductions, then discovers,
    reuses, and consolidates valid macro and parallel contractions.

    \item \textbf{Selection, not exploration volume.}
    We show that RFT also attempts many shortcut-like rewrites, but much of this
    behavior is spurious; RL concentrates exploration into valid reusable
    structure. A finite-group view of GRPO explains how same-prompt contrast can
    suppress spurious passenger actions that RFT clones.

    \item \textbf{Pretraining as procedural substrate.}
    We show that strategy emergence depends on how pretraining organizes
    primitive competence into reduction procedures that RL can later compress
    and reuse.

    \item \textbf{Transfer to a pretrained LLM.}
    With Qwen3-4B on Boolean circuit simplification, RL again solves
    over-budget problems through valid macros, and these macros generalize
    across variable bindings and to a schema never required during training.
\end{enumerate}
\section{Related Work}
\label{sec:related}

\paragraph{RL beyond the base model.}
A central question in RL with verifiable rewards is whether post-training
expands a model's effective capability frontier or mainly reweights behaviors
already latent in the base policy. Yue et al.~\citep{yue2025beyondbase} argue
that RLVR often improves small-$k$ success without expanding the large-$k$
frontier, while support-based analyses emphasize that on-policy RL cannot
amplify zero-probability completions~\citep{wu2025invisibleleash}. In contrast,
ProRL~\citep{liu2025prorl} argues that prolonged RL can make some problems
reachable that were inaccessible under extensive base-model sampling. Closest
to our setting, Yuan et al.~\citep{yuan2025compose} study RL-induced composition
in a controlled synthetic task. Our work studies this question in a setting
where the mechanism is directly observable: the pretraining distribution is
generated by a known grammar, and every emitted solution step can be checked
against that grammar. We return to the comparison with prior composition
experiments in Section~\ref{sec:discussion}.

\paragraph{RL versus imitation-style post-training.}
Rejection and supervised baselines are often strong in reasoning
post-training. STaR bootstraps reasoning from self-generated rationales filtered
by answer correctness~\citep{zelikman2022star}, and recent work shows that
rejection-style or minimalist supervised methods can be competitive with
policy-gradient RL in some RLVR settings~\citep{xiong2025minimalist,chu2025sft}.
We therefore treat rejection fine-tuning as a strong on-policy comparator rather
than a weak baseline: RL and RFT are trained on the same prompts and optimized
separately under the same final-answer supervision.

\paragraph{Process validity, exploration, and pretraining substrate.}
Outcome-only rewards can hide invalid intermediate reasoning
\citep{uesato2022process,lightman2023verify,cobbe2021verifiers}. This is hard to
measure in natural-language tasks, where valid intermediate steps are usually
not enumerable. In our environment, process validity is mechanically checkable,
letting us refine recent discussions of exploration and negative samples in
RLVR~\citep{wang2025hicra,tang2025samplepolarity}. We also connect to work
arguing that post-training depends on capabilities and biases shaped during
pretraining~\citep{zhao2025echo,liu2025r1zero}; here, because pretraining is
controlled, we can ask how the base distribution determines which higher-level
strategies RL can later select, compress, and consolidate.
\section{Rewrite Grammar and Primitive Actions}

\begin{figure}[tbp]
    \centering
    \includegraphics[width=\linewidth,trim={0cm 4cm 0cm 1cm}]
    {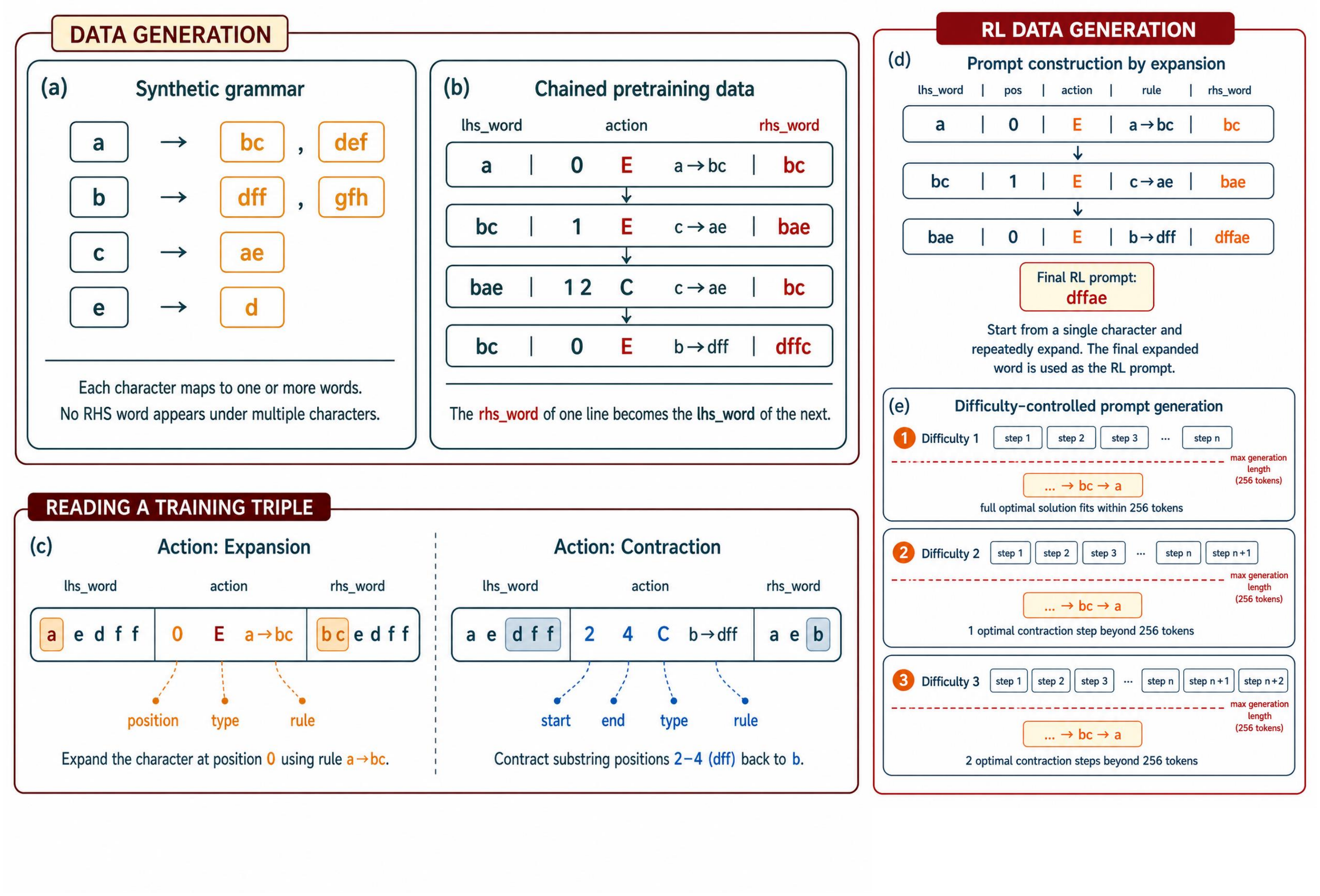}
    \caption{
        Controlled environment and training pipeline.
        \textbf{(a)} Each source character maps to one or more globally unique right-hand-side strings.
        \textbf{(b)} Pretraining data consists of multi-step sequences in which the output of one step
        becomes the input to the next.
        \textbf{(c)} Each training triple encodes a single primitive expansion or contraction.
        \textbf{(d)} RL prompts are constructed by repeated forward expansion from a target symbol.
        \textbf{(e)} Difficulty levels are defined by how many optimal contraction steps lie beyond the
        model's 256-token generation budget.
    }
    \label{fig:schematic}
\end{figure}

The grammar is defined over an alphabet $\mathcal{A}$, where each source symbol
maps to one or more globally unique multi-character right-hand-side strings
(Figure~\ref{fig:schematic}a). Global uniqueness makes primitive contraction
unambiguous: any substring matching a right-hand side maps back to exactly one
source symbol. The environment supports two inverse primitive operations:
\emph{expansion}, which replaces a symbol by one of its right-hand sides, and
\emph{contraction}, which replaces a matching substring by its source symbol.
Each step is serialized as
\[
\texttt{lhs\_word} \mid \texttt{action} \mid \texttt{rhs\_word},
\]
where the action specifies a position and rule for expansion, or a span and
rule for contraction (Figure~\ref{fig:schematic}c).

Because the primitive rule set is known, every generated rewrite can be audited
against the grammar: it can be a primitive rule application, a valid
non-primitive shortcut, or an invalid rewrite. This is what makes
trajectory-level strategy analysis possible.

\subsection{Pretraining on Primitive Rewrite Chains}

The model is pretrained from scratch on chained rewrite trajectories. Starting
from an initial word, the data generator repeatedly samples a valid primitive
expansion or contraction, appends the corresponding serialized triple, and uses
the resulting word as the input to the next step. Thus, pretraining exposes the
model to local grammar-consistent rewrite dynamics, but not to the later
non-primitive shortcuts analyzed in this paper.

We control the pretraining distribution with a contraction weight $\rho$, which
locally upweights contraction moves relative to expansion moves whenever both
are available. Thus $\rho$ affects the realized contraction frequency and the
amount of contraction chaining, but does not directly specify either statistic.
Unless otherwise stated, we use $\rho=4$. Section~\ref{sec:pretraining_substrate} later ablates this parameter to test whether
properties of the pretraining distribution affect which strategies RL can
unlock. Full grammar-generation and optimization details are given in
Appendix~\ref{app:training}.

\subsection{Goal-Directed Contraction Task}

After pretraining, the model is post-trained on a goal-directed contraction
task. We sample a target symbol $c^\star$ and repeatedly apply primitive
expansions to produce a prompt word $w_{\mathrm{prompt}}$. At post-training
time, the model sees only $w_{\mathrm{prompt}}$ and must generate a contraction
trajectory whose final emitted symbol is $c^\star$.

Difficulty is defined by the length of the optimal primitive contraction
solution under the serialized action format. Difficulty~1 problems have an
optimal primitive solution that fits within the 256-token generation budget.
Difficulty~2 problems require one additional primitive contraction step beyond
the budget, and Difficulty~3 problems require two additional steps. The RL
training mixture contains 60\% Difficulty~1, 20\% Difficulty~2, and 20\%
Difficulty~3 examples. For evaluation, we use held-out problems spanning
Difficulties~1--5, including buckets beyond the training distribution.

This difficulty definition is important: higher difficulty does not mean the
grammar is different, but that the straightforward primitive contraction path is
too long for the generation budget. Solving harder buckets therefore requires
either more efficient use of primitive rules or shorter non-primitive solution
paths.

\subsection{RL and RFT Post-Training}

The post-training reward is binary and outcome-only: it is $1$ if the
trajectory consists of well-formed action triples in the format above and
the final emitted symbol equals the target $c^\star$, and $0$ otherwise.
There is no supervision on intermediate rewrites — only the format of the
output and the identity of the final symbol matter. Any valid non-primitive
strategies that emerge must therefore arise from post-training dynamics
rather than from direct reward shaping. This mirrors RLVR on math reasoning,
where final-answer correctness is the only training
signal~\citep{shao2024deepseekmath}.

Our RL method is Group Relative Policy Optimization
(GRPO)~\citep{shao2024deepseekmath}. As an imitation baseline, we use
rejection fine-tuning (RFT), which retains successful on-policy rollouts and
performs next-token prediction on them. RFT is therefore a strong comparator:
any RL-vs-RFT gap cannot be attributed to data or prompt mismatch. Full GRPO
and RFT details are in Appendix~\ref{app:grpo}.

\section{RL Expands the Held-Out Capability Frontier}
\label{sec:frontier}

We first ask whether RL expands what the model can solve relative to the
pretrained base policy, or merely reweights solutions already accessible under
it. Our setting lets us answer this in a finite-budget, fully observable form:
we can measure base-model reach under larger sampling budgets and inspect the
structure of the resulting trajectories. We evaluate on the held-out difficulty
buckets defined above, where higher difficulty corresponds to ordinary
grammar-rule solutions becoming increasingly incompatible with the 256-token
generation budget.

\begin{figure}[!htbp]
    \centering
    \includegraphics[
        width=0.8\linewidth,
        trim={0cm 0.0cm 0cm 0.0cm},
        clip
    ]{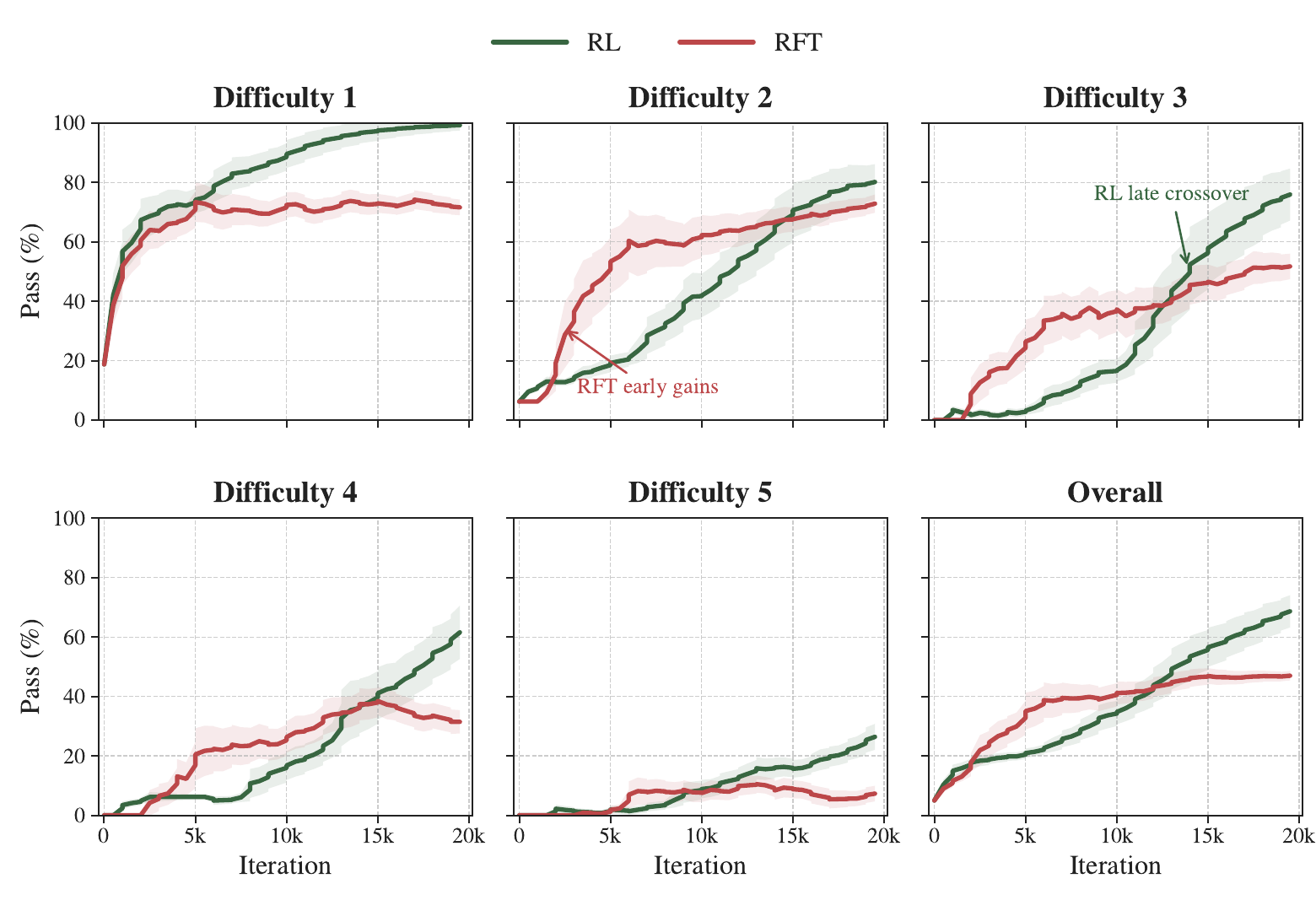}
    \caption{
        RL exhibits a delayed crossover over RFT, with the largest advantage on harder
        held-out problems.
        Pass@16 on contraction problems across five difficulty levels and overall. Mean over three seeds.
        RFT improves rapidly early in training but plateaus, while RL improves more
        gradually and continues climbing through the end of training.
    }
    \label{fig:pass16}
\end{figure}

Figure~\ref{fig:pass16} shows pass@16 during post-training. RFT improves
faster at the beginning of training, especially on easier buckets, indicating
that on-policy imitation can amplify solutions the pretrained model already
sometimes produces~\citep{xiong2025minimalist}. RFT then plateaus, while RL
improves more slowly at first but continues climbing after RFT saturates, with
the largest gap appearing on the hardest buckets, where primitive contraction
paths are least compatible with the generation budget.

To check whether the harder buckets are merely under-sampled by the base model,
we evaluate the pretrained policy at much larger sampling budgets
(Table~\ref{tab:base_passk}). Even at pass@1024, base-model success remains
sparse and concentrated in the easiest over-budget buckets, with Buckets~4--5
remaining at zero. Thus, under the tested finite budgets, complete hard-bucket
solutions are not practically available from the pretrained model, whereas RL
makes them reliably accessible at pass@16.

How does RL cross this frontier? Over-budget problems require shortcuts, which
could be invalid rewrites that exploit the outcome-only reward or valid
compressed procedures; the next section distinguishes these by auditing the
generated trajectories.

\begin{table}[t]
\centering
\small
\caption{
Base-model pass@$k$ on held-out over-budget problems.
Overall aggregates across 80 problems; each difficulty bucket contains 16.
}
\label{tab:base_passk}
\begin{tabular}{lcccccc}
\toprule
$k$ & \textbf{Overall} & \textbf{Difficulty 1} & \textbf{Difficulty 2} & \textbf{Difficulty 3} & \textbf{Difficulty 4} & \textbf{Difficulty 5} \\
\midrule
64   & 11.25\% & 50.00\% &  6.25\% & 0.00\% & 0.00\% & 0.00\% \\
256  & 13.75\% & 50.00\% & 12.50\% & 6.25\% & 0.00\% & 0.00\% \\
1024 & 17.50\% & 62.50\% & 18.75\% & 6.25\% & 0.00\% & 0.00\% \\
\bottomrule
\end{tabular}
\end{table}

\section{How the Model Goes Beyond the Base Model}

Because the grammar is fully known, every generated rewrite in a successful
trajectory can be audited by comparing its left- and right-hand sides. A
\textbf{primitive contraction} replaces the right-hand side of an original
grammar rule with its source symbol; for example, \texttt{dff} contracts to
\texttt{b} when the grammar contains $b \rightarrow \texttt{dff}$. A
\textbf{macro contraction} is a non-primitive rewrite equivalent to a sequence
of primitive contractions, such as \texttt{bae} $\rightarrow$ \texttt{a} by
composing \texttt{ae} $\rightarrow$ \texttt{c} and \texttt{bc} $\rightarrow$
\texttt{a}. A \textbf{parallel contraction} applies independent primitive
contractions simultaneously, such as \texttt{dffae} $\rightarrow$ \texttt{bc}. Together these two action types — sequential and parallel — instantiate the procedural chunks introduced in Section 1: compressed multi-step procedures that can stand in for chains or combinations of primitive operations.
Any non-primitive rewrite that is neither macro nor parallel is classified as
\textbf{spurious}.

\subsection{Dynamics of Strategy Emergence}

Recent RL post-training work reports phase-like changes in reasoning behavior:
models first improve lower-level solution execution and later exhibit reflection,
alternative approaches, or higher-level strategic behavior
\citep{guo2025deepseekr1,wang2025hicra,cheng2025exploration}. In natural-language
settings, these phases are hard to audit because intermediate reasoning steps
are not mechanically checkable. Our grammar gives a trace-level analogue: each
generated rewrite can be classified against the known rule system, letting us
measure which kind of procedure becomes more common during training.

\begin{figure}[!htbp]
    \centering
    \includegraphics[
        width=0.8\linewidth,
        trim={0cm 0cm 0cm 0cm},
        clip
    ]{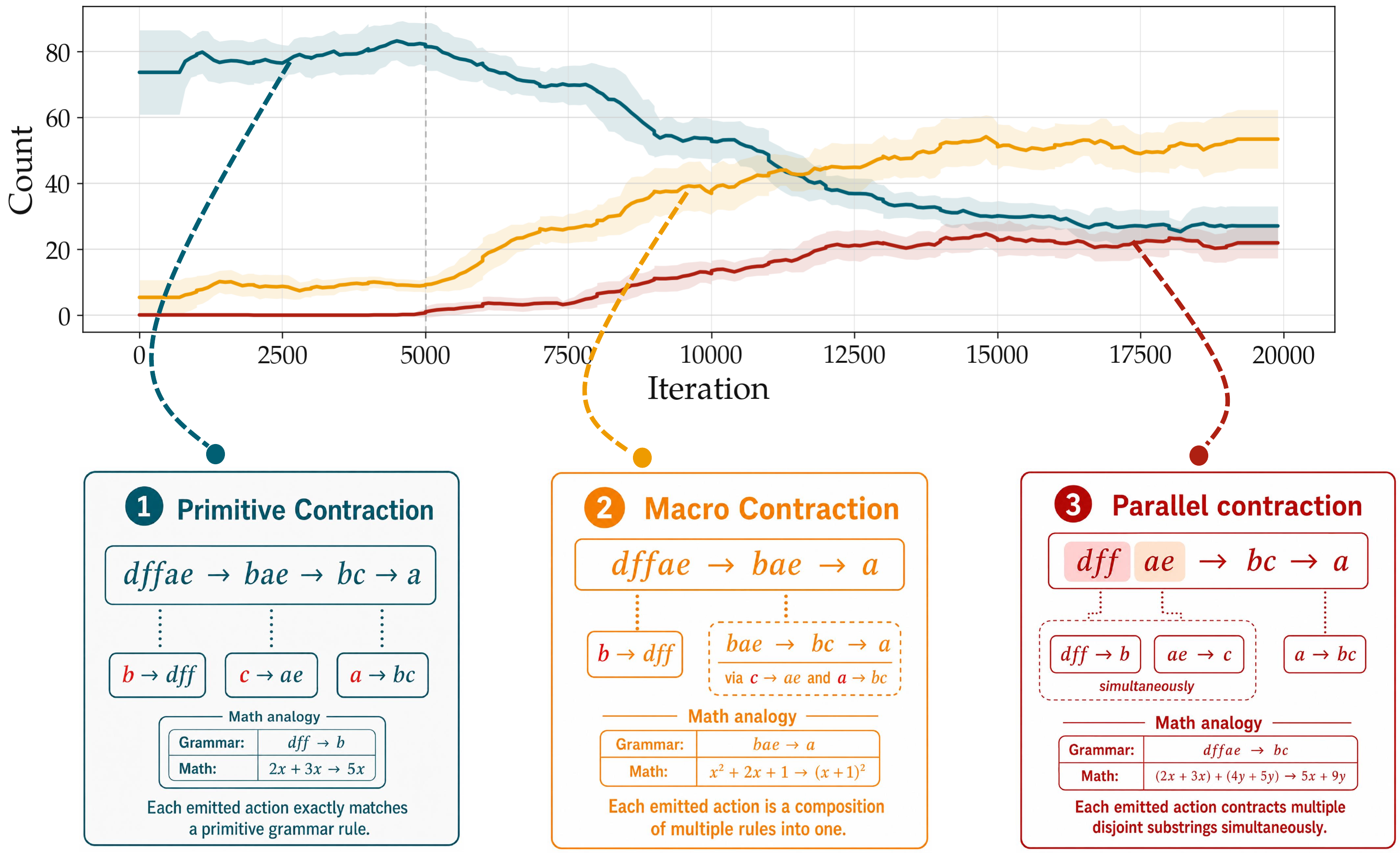}
        \caption{
    RL undergoes a delayed transition from primitive contractions to compressed
    valid shortcuts.
    Macro contractions accelerate and overtake primitive contractions around
    iteration~12{,}500; parallel contractions emerge later.
    Bottom examples show the structural analogy to algebraic simplification:
    primitive contractions correspond to local simplifications, macro
    contractions to reusable multi-step maneuvers, and parallel contractions to
    independent simplifications applied simultaneously.
}
    
    \label{fig:strategy_phases}
\end{figure}

Figure~\ref{fig:strategy_phases} shows a delayed transition. Early successful
trajectories are dominated by primitive contractions. Around
iteration~5{,}000, macro contractions begin to rise and overtake primitive
contractions by roughly iteration~12{,}500, while parallel contractions
emerge later and grow steadily.

Thus, RL first strengthens primitive reduction behavior and only later shifts
toward compressed valid procedures that were never present as primitive training
actions. This trace-level phase transition explains the late gains in
Figure~\ref{fig:pass16}: on the hardest buckets, primitive paths are least
compatible with the generation budget, so valid compressed rewrites make
previously over-budget solutions reachable.

\subsection{Discovery and Consolidation of Macro Rules}

The previous subsection showed that macro contractions become increasingly
important during RL training. We now ask whether this increase reflects
scattered one-off shortcuts, or whether RL builds a reusable repertoire of
macro rules. We distinguish between a \emph{macro action}, a single occurrence
of a macro contraction in a trajectory, and a \emph{macro rule}, the underlying
rewrite pattern that may be used repeatedly across trajectories. We measure
\emph{discovery} as the first appearance of a macro rule, and
\emph{consolidation} as the later reuse of previously discovered macro rules.

\begin{figure}[tbp]
    \centering
    \includegraphics[width=0.8\linewidth,
    trim={0cm 0cm 0cm 1.5cm},
        clip]
    {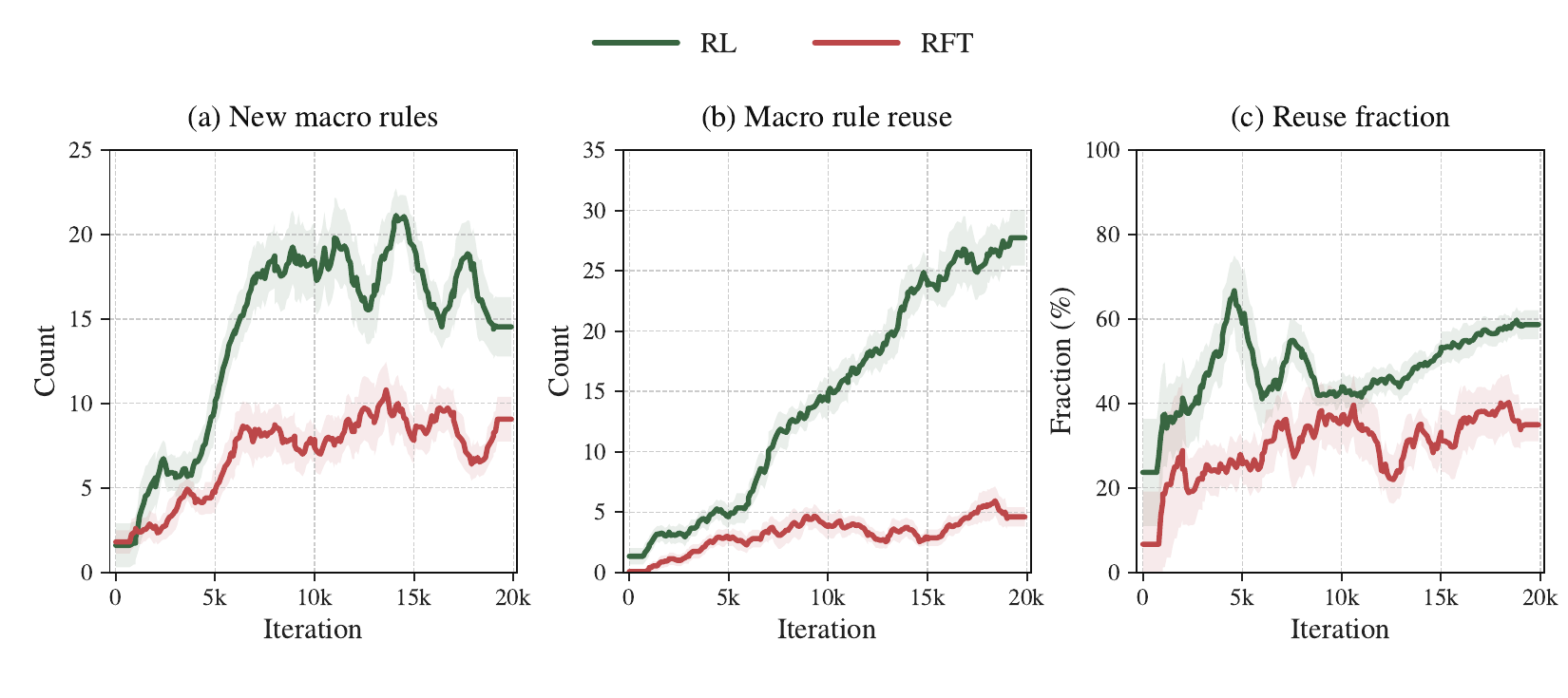}
    \caption{
RL discovers and consolidates reusable macro rules.
\textbf{(a)} Number of newly discovered macro rules at each checkpoint.
\textbf{(b)} Reuse count of previously discovered macro rules.
\textbf{(c)} Reuse fraction among all macro actions.
RL continues discovering new macro rules while increasingly reusing earlier ones, indicating consolidation into a stable repertoire.
}
    \label{fig:discovery_consolidation}
\end{figure}

Figure~\ref{fig:discovery_consolidation} shows that RL's macro behavior is not
just an accumulation of isolated shortcuts. RL continues discovering new macro
rules throughout training, with discovery peaking around iteration~10{,}000.
Reuse rises in parallel: by the end of training, most macro actions emitted by
the RL model are reapplications of rules observed at earlier checkpoints. RFT
also discovers some macro rules, but at a lower rate and with weaker reuse: it
finds shortcuts, but does not consolidate them into an equally stable repertoire.

This consolidation is the bridge from discovery to capability. A macro
contraction compresses a sequence of primitive contractions into a single
emitted rewrite, behaving like an implicit macro-action without any option set
or high-level controller being provided
\citep{sutton1999between,konidaris2009skillchaining}. The abstractions emerge
inside the sequence model and are visible only because the trace can be audited.
Thus RL accumulates capability not by relying on a fresh lucky shortcut each
time, but by stabilizing discovered shortcuts into reusable policy structure
that can be deployed on later problems.
\section{Structured Versus Unstructured Exploration}
\label{sec:structured_exploration}
\begin{figure}[tbp]
    \centering
    \includegraphics[
        width=0.8\linewidth,
        clip,trim={0cm 0.0cm 0cm 0cm}
    ]{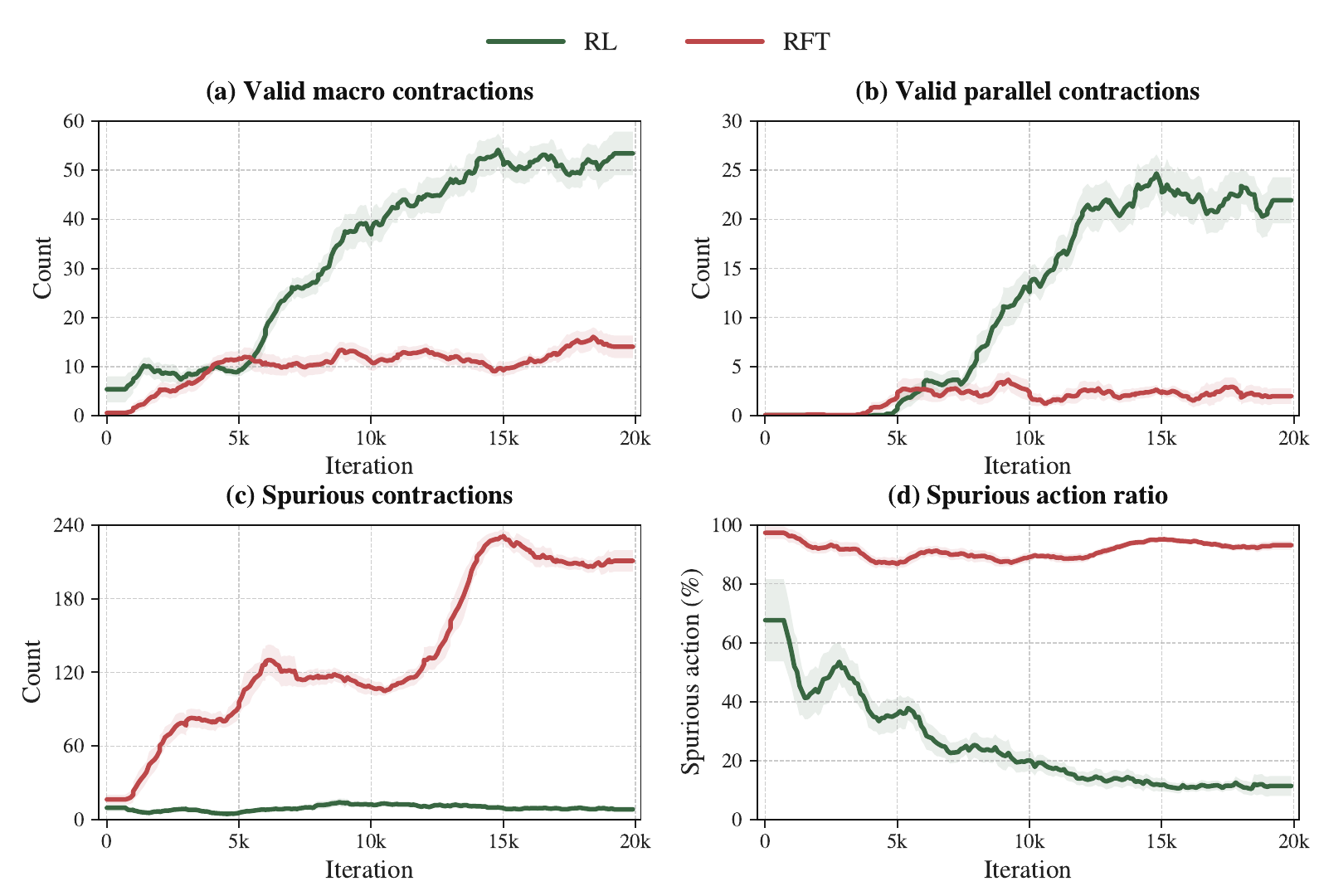}
    \caption{
RL discovers valid non-primitive strategies while suppressing spurious shortcuts.
\textbf{(a,b)} Counts of valid macro and parallel contractions.
\textbf{(c)} Count of spurious contractions.
\textbf{(d)} Fraction of non-primitive contractions that are spurious.
RFT generates many non-primitive actions, but they are mostly invalid; RL produces fewer spurious rewrites and increasingly concentrates behavior into valid macro and parallel contractions.
}
    \label{fig:strategy_classes_four_panel}
\end{figure}

The previous section established that RL discovers and consolidates valid macro
rules more effectively than RFT. A natural alternative explanation is that RL
simply explores more: perhaps it generates more non-primitive actions in total
and therefore finds more valid shortcuts by chance.
Figure~\ref{fig:strategy_classes_four_panel} rules this out. Panels~(a) and~(b) show that both methods
generate valid non-primitive behavior: RFT discovers macro contractions and
parallel contractions, but RL discovers many more of both, especially later in
training. Panels~(c) and~(d) show the key difference. RFT accumulates many more
spurious contractions, and its spurious-action ratio remains high throughout
training. RL instead keeps spurious contractions low and shifts an increasing
fraction of its non-primitive behavior toward valid structure. The difference is
therefore not raw non-primitive volume, but selectivity.

A finite-group view of GRPO explains how this selectivity can arise. Consider
any binary trajectory feature $F(\tau)$, such as the presence of a particular
macro contraction, parallel contraction, or spurious contraction. For a prompt
group of size $G$, let $n$ be the number of successful completions. Let $k^+$
be the number of successful completions containing $F$, and let $k^-$ be the
number of failed completions containing $F$. With binary group-normalized
rewards, the feature-level GRPO update is proportional, up to a positive scale,
to
\definecolor{successgreen}{RGB}{36,130,90}
\definecolor{failred}{RGB}{170,60,60}

\[
\underbrace{\textcolor{successgreen}{\frac{k^+}{n}}}_{\text{frequency among successes}}
-
\underbrace{\textcolor{failred}{\frac{k^-}{G-n}}}_{\text{frequency among failures}}
\]
Thus GRPO increases features that are more common in successful completions than
failed completions from the same prompt, and suppresses features enriched in
failures. Appendix~\ref{app:feature_selection} derives this expression from the
group-normalized advantages.

RFT uses the sampled rollouts differently. It discards failed completions,
retains only successful trajectories, and then performs supervised learning on
the full accepted sequence. As a result, if a successful trajectory contains a
spurious rewrite alongside useful steps, that rewrite is treated as part of the
target sequence. GRPO does not identify which individual action caused success,
but it does compare successes against failures from the same prompt. This
within-group contrast gives downward pressure to action patterns that appear
frequently in failed rollouts, even if they also occur inside some successful
ones.

This sharpens recent accounts of negative samples in RLVR. Prior work has argued
that negative samples can help maintain exploration, preserve diversity, or
prevent premature collapse \citep{wang2025hicra,tang2025samplepolarity}. Our setting shows that this framing misses the actual mechanism. Because every emitted action can be
classified, we can see that RFT already produces abundant non-primitive behavior;
the bottleneck is not exploration volume. The key difference is that RL filters
this explored space: it amplifies non-primitive actions that are valid and
reward-aligned, while suppressing spurious departures from the grammar. Thus the
role of negative samples is not merely to keep the policy diverse, but to make
exploration structurally selective.

\section{Pretraining as the Substrate for Strategy Emergence}
\label{sec:pretraining_substrate}

The previous sections show that RL eventually discovers and consolidates
valid non-primitive strategies. We now ask what the base model must already
provide --- not whether pretraining contains macro or parallel strategies, but
whether it organizes primitive competence into reduction procedures that RL
can later compress.

\begin{figure}[!htbp]
    \centering
    \includegraphics[width=0.85\linewidth]{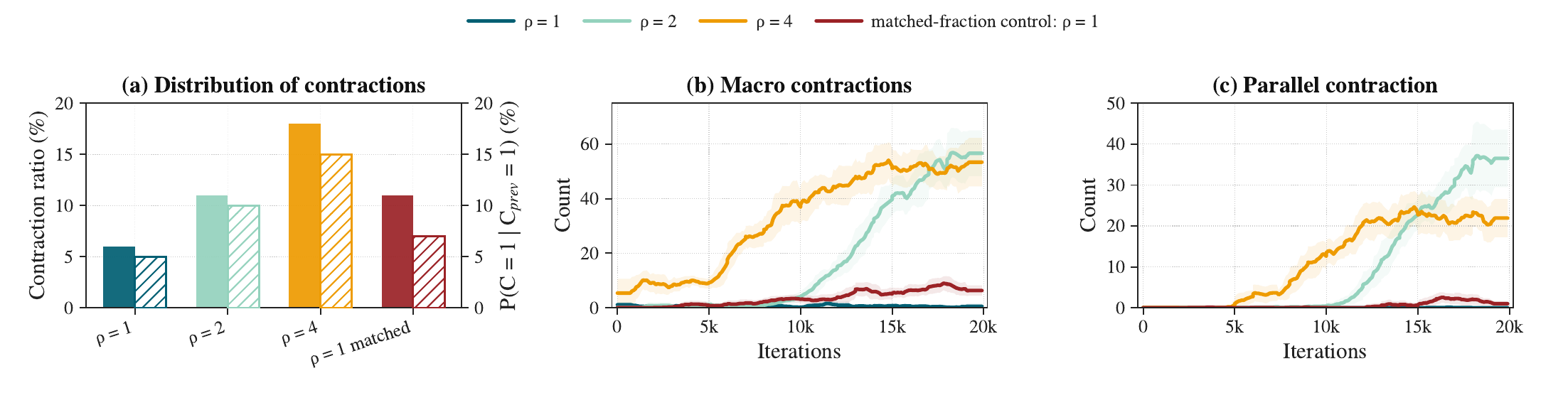}
    \caption{
        Strategy emergence depends on local contraction structure, not
        overall contraction frequency.
        \textbf{(a)} Solid bars (left axis): overall contraction rate.
        Hatched bars (right axis): contraction-after-contraction probability
        $P(C_t=1 \mid C_{t-1}=1)$. The matched-fraction control reaches a
        comparable overall rate to $\rho=2$ but has substantially less
        chaining.
        \textbf{(b,c)} Macro and parallel contraction counts during RL
        post-training. Higher $\rho$ produces earlier and stronger
        non-primitive strategy emergence; the matched-fraction control fails
        despite comparable marginal exposure.
    }
    \label{fig:rho}
\end{figure}

We vary how strongly contractions are locally upweighted during pretraining
via a parameter $\rho$ (Appendix~\ref{app:rho_pretraining}). Increasing $\rho$
has two coupled effects: contractions occur more often overall, and once a
reducible state is reached, the next step is more likely to also be a
contraction --- producing chained primitive reductions rather than isolated
contraction events. To separate these effects, we include a matched-fraction
control: a low-$\rho$ run with marginal contraction rate adjusted to match
$\rho=2$ but without the local upweighting that produces chaining.
Figure~\ref{fig:rho}a confirms the design: the control matches $\rho=2$ on
overall contraction rate (solid bars) but has less chaining
(hatched bars, $P(C_t=1\mid C_{t-1}=1)$).

Figures~\ref{fig:rho}b,c show that this distinction matters. Low-$\rho$
pretraining never reliably enters the macro or parallel regime; higher $\rho$
produces earlier and stronger valid non-primitive strategies; the
matched-fraction control fails to recover the $\rho=2$ dynamics despite
comparable marginal contraction exposure, behaving similarly to $\rho=1$.
What matters is not how often the model sees contractions, but whether they
appear as sustained reduction procedures.

RL does not upweight a fully-formed latent strategy already present in the
base model. Pretraining supplies weak procedural ingredients --- primitive
contractions, reduction states, and short contraction chains --- that RL
later selects, compresses, and consolidates into reusable higher-level
strategies.

\section{RL Discovers Transferable Macros in a Pretrained LLM}
\label{sec:boolean}

The grammar isolates the mechanism in a small model trained from scratch, but
its symbols are arbitrary, so each macro is a fixed string and the grammar
cannot rule out that RL memorizes a set of rewarded rewrites. We now ask whether
the mechanism holds in a pretrained LLM two orders of magnitude larger, and
whether its macros are reusable operations rather than memorized strings. We
find that it does:
RL again crosses the budget frontier through valid macros, and these macros
generalize across variable bindings and to a schema never required during
training.

\begin{figure}[t]
\centering
\textbf{(a)} Same structure in both settings\\[3pt]
{\small
\begin{tabular}{@{}llll@{}}
\toprule
 & Input & Primitive steps & Macro step \\
\midrule
Grammar & \texttt{bae} &
  $\xrightarrow{\texttt{ae}\rightarrow\texttt{c}}$ \texttt{bc}
  $\xrightarrow{\texttt{bc}\rightarrow\texttt{a}}$ \texttt{a} &
  $\rightarrow$ \texttt{a} \\[4pt]
Boolean & $g_1 = B \oplus B,\ g_2 = A \oplus g_1$ &
  $\xrightarrow{\text{elim.\ } g_1}$ $g_1 = 0$
  $\xrightarrow{\text{elim.\ } g_2}$ $g_2 = A$ &
  $\rightarrow$ $g_2 = A$ \\
\bottomrule
\end{tabular}}

\vspace{3pt}
\begin{minipage}[t]{0.42\linewidth}
\centering
\textbf{(b)} Frontier\\[2pt]
\begin{tikzpicture}
\begin{axis}[
  width=\linewidth, height=3.4cm,
  ybar, bar width=6.5pt,
  ymin=0, ymax=46, ytick={0,20,40},
  ylabel={Solved (of 40)},
  xmin=-0.6, xmax=4.6, xtick={0,1,2,3,4},
  xticklabels={D0,D1,D2,D3,D4},
  tick label style={font=\scriptsize}, label style={font=\scriptsize},
  axis lines*=left, ymajorgrids, grid style={black!10},
  every node near coord/.append style={font=\tiny, text=black!75},
  nodes near coords,
  legend style={font=\scriptsize, draw=none, fill=none, at={(0.5,1.03)}, anchor=south, legend columns=2, /tikz/every even column/.append style={column sep=6pt}},
  legend image code/.code={\draw[#1, draw=none] (0cm,-0.08cm) rectangle (0.25cm,0.12cm);},
  clip=false,
]
\fill[black!4] (axis cs:0.5,0) rectangle (axis cs:4.6,46);
\node[font=\scriptsize\itshape, text=black!55, anchor=north] at (axis cs:2.55,45) {over budget};
\addplot[fill=sftgray, draw=none] coordinates {(0,39) (1,3) (2,1) (3,1) (4,0)};
\addplot[fill=rlgreen, draw=none] coordinates {(0,37) (1,27) (2,20) (3,6) (4,4)};
\legend{SFT (pass@1024), GRPO (pass@16)}
\end{axis}
\end{tikzpicture}
\end{minipage}\hfill
\begin{minipage}[t]{0.56\linewidth}
\centering
\textbf{(c)} Schema transfer\\[2pt]
\begin{tikzpicture}
\begin{axis}[
  width=0.62\linewidth, height=3.4cm, scale only axis=false,
  xbar, bar width=6.5pt,
  xmin=0, xmax=100, xtick={0,50,100},
  xlabel={Correct applications (\%)},
  ymin=-0.6, ymax=4.6, y dir=reverse,
  ytick={1,2,4},
  yticklabels={$X \oplus (Y \oplus Y) \rightarrow X$, $X \land (Y \oplus \lnot Y) \rightarrow X$, $X \lor (Y \oplus Y) \rightarrow X$},
  tick label style={font=\scriptsize}, label style={font=\scriptsize},
  axis lines*=left, xmajorgrids, grid style={black!10},
  clip=false,
]
\node[font=\scriptsize\itshape, text=black!60, anchor=west] at (axis cs:1,0) {required during RL};
\node[font=\scriptsize\itshape, text=black!60, anchor=west] at (axis cs:1,3) {never required during training};
\draw[black!25, dashed] (axis cs:0,2.5) -- (axis cs:100,2.5);
\addplot[fill=rlgreen, draw=none] coordinates {(100,1) (100,2) (77.8,4)};
\node[font=\tiny, text=black!75, anchor=west] at (axis cs:100,1) {20/20};
\node[font=\tiny, text=black!75, anchor=west] at (axis cs:100,2) {22/22};
\node[font=\tiny, text=black!75, anchor=west] at (axis cs:77.8,4) {7/9};
\end{axis}
\end{tikzpicture}
\end{minipage}
\caption{
In a pretrained LLM, RL discovers macros that generalize as operations.
\textbf{(a)} The Boolean task mirrors the grammar: a primitive step applies one
rule or eliminates one gate, and a macro collapses several primitive steps into
one. Gates $g_1, g_2$ are computed from inputs $A, B$ (shown with readable
names); this macro is an instance of the first schema in (c).
\textbf{(b)} Held-out circuits solved per difficulty bucket by Qwen3-4B.
\textbf{(c)} Evaluation circuits on which GRPO correctly applies each macro
schema; $X$ and $Y$ stand for any subexpression.
}
\label{fig:boolean}
\end{figure}

\paragraph{Setup.}
We post-train Qwen3-4B~\citep{yang2025qwen3} on Boolean circuit
simplification. A circuit is a sequence of gate assignments; a primitive step
eliminates one gate, and a macro collapses several eliminations into one step
(Figure~\ref{fig:boolean}a), mirroring primitive and macro contractions.
Because a pretrained model may already know such simplifications, gates are
opaque tokens whose meaning is taught only by an initial SFT stage that
demonstrates single-gate eliminations but none of the macros analyzed below.
GRPO then trains from this checkpoint with a final-answer reward. As in the
grammar, difficulty is set by the generation budget (D0 admits a primitive
solution within budget; D1--D4 do not), and we audit every generated step.
Full details are in Appendix~\ref{app:boolean}.

\paragraph{RL crosses the frontier with valid macros.}
The SFT policy solves in-budget problems (39/40 on D0) but rarely solves
over-budget ones even at pass@1024 (Figure~\ref{fig:boolean}b). GRPO solves 27,
20, 6, and 4 of 40 over-budget problems at pass@16, a $64\times$ smaller
sampling budget, and every over-budget success uses at least one valid macro.
As in the grammar, RL makes valid compressed procedures accessible that the
starting policy does not practically produce.

\paragraph{Macros generalize as operations.}
Figure~\ref{fig:boolean}c probes two levels of generalization. First, since
variable names are resampled, no macro recurs as the same string; the two
schemas required during RL are nevertheless applied correctly on 20/20 and
22/22 evaluation circuits with distinct bindings, so RL has learned the
operation rather than memorizing particular strings. Second, $X \lor (Y \oplus Y)
\rightarrow X$ was never demonstrated or required during training, yet GRPO
applies it correctly on 7/9 evaluation circuits, while the SFT policy applies
no undemonstrated schema. This schema shares the structure of the trained ones,
collapsing a subexpression that evaluates to a constant, so RL extends a learned
operation to a structurally related case it was never trained on.

Together, these results show that the chunking mechanism identified in the
grammar is not an artifact of small models trained from scratch. In a
pretrained LLM, outcome-only RL constructs valid compressed operations that the
starting policy does not practically produce, and these operations reach beyond
the cases that training required.

\section{Discussion}
\label{sec:discussion}

The controlled grammar turns the ``beyond the base model'' question into an
inspectable one and exposes a concrete mechanism: RL strengthens primitive
reductions, then forms and consolidates valid compressed procedures. This is a
form of procedural chunking, long studied in skill acquisition such as chess
expertise~\citep{chase1973perception}; in mathematics, multi-step derivations
similarly become reusable lemmas or standard maneuvers. Macro and parallel
contractions are measurable instances of such chunks.

Our results challenge the view that RL merely reweights complete reasoning paths
already present in the base distribution~\citep{yue2025beyondbase}. In this
environment, the complete hard-bucket solution is not practically present in the
base model: even at pass@1024, the pretrained policy gets $0\%$ on
Buckets~4--5, while RL later solves them at pass@16. What is present are
ingredients---primitive contractions and weakly supported local reduction
chains---which RL assembles into complete procedures. Thus RL goes beyond the
base model not by
inventing from nothing~\citep{wu2025invisibleleash}, but by constructing usable
strategies from weakly supported components. The base model provides the
ingredients; RL builds the procedures.

This clarifies the comparison to Yuan et al.~\citep{yuan2025compose}. Their
setting starts from a large pretrained model, so RL-induced composition is hard
to separate from amplification of compositional patterns already supported by
pretraining. It also leaves open which pretraining structures make composition
reachable. Their task evaluates direct input-output transformations, whereas
reasoning models often rely on extended traces where the procedure itself
matters. Our setting addresses these confounds: pretraining is controlled and
ablated, the primitive rule set is known, and the model emits the full rewrite
trajectory.

The pretraining ablation adds that the relevant substrate is not exposure to
primitive facts alone but their organization into reduction procedures:
pretraining supplies the procedural ingredients, and RL performs the selection
and chunking that turn them into stable strategies. More broadly, while synthetic by
design, the grammar exposes quantities normally hidden in LLM post-training, and these results
suggest that RL post-training is not only reward-driven reweighting; it can also
organize weak procedural ingredients into reusable reasoning structure.

\bibliographystyle{plainnat}
\bibliography{references}
\section*{NeurIPS Paper Checklist}

\begin{enumerate}

\item {\bf Claims}
    \item[] Question: Do the main claims made in the abstract and introduction accurately reflect the paper's contributions and scope?
    \item[] Answer: \answerYes{}.
    \item[] Justification: The abstract and introduction frame the work as a controlled study of RL post-training in a synthetic rewrite grammar, complemented by a controlled experiment with a pretrained LLM (Section~\ref{sec:boolean}), and the claims are limited to the contributions and empirical evidence presented in the paper.

\item {\bf Limitations}
    \item[] Question: Does the paper discuss the limitations of the work performed by the authors?
    \item[] Answer: \answerYes{}.
    \item[] Justification: The paper presents the grammar as a deliberately synthetic, controlled setting, and Section~\ref{sec:boolean} tests how far its central result extends to a pretrained LLM; the results are not claimed to universally characterize large language models.

\item {\bf Theory assumptions and proofs}
    \item[] Question: For each theoretical result, does the paper provide the full set of assumptions and a complete (and correct) proof?
    \item[] Answer: \answerYes{}.
    \item[] Justification: The paper's only theoretical result is the finite-group feature-contrast expression in Appendix C, derived in full under its stated simplifications (ignoring token-level constants, clipping, and the KL term). It is a directional argument for how GRPO's same-prompt contrast acts on trajectory features, not a model of the full GRPO training dynamics.

\item {\bf Experimental result reproducibility}
    \item[] Question: Does the paper fully disclose all the information needed to reproduce the main experimental results of the paper to the extent that it affects the main claims and/or conclusions of the paper (regardless of whether the code and data are provided or not)?
    \item[] Answer: \answerYes{}.
    \item[] Justification: The paper specifies the grammar generation, data generation, model architecture, optimization setup, reward, GRPO/RFT procedures, and evaluation protocol needed to reproduce the main experiments; Appendix~\ref{app:boolean} describes the Boolean-circuit experiment with Qwen3-4B.

\item {\bf Open access to data and code}
    \item[] Question: Does the paper provide open access to the data and code, with sufficient instructions to faithfully reproduce the main experimental results, as described in supplemental material?
    \item[] Answer: \answerNo{}.
    \item[] Justification: The data are synthetically generated and the paper provides the generation and experimental details needed for reproducibility, but the code release is planned after acceptance rather than during anonymous submission.

\item {\bf Experimental setting/details}
    \item[] Question: Does the paper specify all the training and test details (e.g., data splits, hyperparameters, how they were chosen, type of optimizer) necessary to understand the results?
    \item[] Answer: \answerYes{}.
    \item[] Justification: The paper specifies the grammar, prompt construction, difficulty buckets, model architecture, optimizer, reward, GRPO/RFT settings, and evaluation protocol; Appendix~\ref{app:boolean} gives the corresponding details for the Boolean-circuit experiment.

\item {\bf Experiment statistical significance}
    \item[] Question: Does the paper report error bars suitably and correctly defined or other appropriate information about the statistical significance of the experiments?
    \item[] Answer: \answerYes{}.
    \item[] Justification: The main grammar post-training comparisons are reported over three seeds, with uncertainty shown in the corresponding plots; for the Boolean-circuit experiment we report solve counts on 40 held-out circuits per difficulty bucket.

\item {\bf Experiments compute resources}
    \item[] Question: For each experiment, does the paper provide sufficient information on the computer resources (type of compute workers, memory, time of execution) needed to reproduce the experiments?
    \item[] Answer: \answerYes{}.
    \item[] Justification: The appendix reports the compute hardware used for the experiments, including that the main runs were performed on 4$\times$H100 GPUs.

\item {\bf Code of ethics}
    \item[] Question: Does the research conducted in the paper conform, in every respect, with the NeurIPS Code of Ethics \url{https://neurips.cc/public/EthicsGuidelines}?
    \item[] Answer: \answerYes{}.
    \item[] Justification: The work uses synthetic grammar-generated data and synthetic Boolean circuits, and does not involve human subjects, private data, or scraped datasets.

\item {\bf Broader impacts}
    \item[] Question: Does the paper discuss both potential positive societal impacts and negative societal impacts of the work performed?
    \item[] Answer: \answerYes{}.
    \item[] Justification: The work is a controlled synthetic study with no direct deployment, but it may positively improve understanding of RL post-training and may indirectly inform capability-building methods.

\item {\bf Safeguards}
    \item[] Question: Does the paper describe safeguards that have been put in place for responsible release of data or models that have a high risk for misuse (e.g., pre-trained language models, image generators, or scraped datasets)?
    \item[] Answer: \answerNA{}.
    \item[] Justification: The paper does not release high-risk models or scraped datasets; the experiments use synthetic grammar-generated data in a controlled environment.

\item {\bf Licenses for existing assets}
    \item[] Question: Are the creators or original owners of assets (e.g., code, data, models), used in the paper, properly credited and are the license and terms of use explicitly mentioned and properly respected?
    \item[] Answer: \answerYes{}.
    \item[] Justification: We use the open-weight Qwen3-4B model~\citep{yang2025qwen3}, which is credited in the paper and released under the Apache 2.0 license, and we use it in accordance with its terms. All experimental data are synthetically generated.

\item {\bf New assets}
    \item[] Question: Are new assets introduced in the paper well documented and is the documentation provided alongside the assets?
    \item[] Answer: \answerNA{}.
    \item[] Justification: The paper does not release new assets during anonymous submission; the synthetic environment and data-generation procedure are documented in the paper and appendix.

\item {\bf Crowdsourcing and research with human subjects}
    \item[] Question: For crowdsourcing experiments and research with human subjects, does the paper include the full text of instructions given to participants and screenshots, if applicable, as well as details about compensation (if any)?
    \item[] Answer: \answerNA{}.
    \item[] Justification: The paper does not involve crowdsourcing, human annotators, user studies, or research with human subjects.

\item {\bf Institutional review board (IRB) approvals or equivalent for research with human subjects}
    \item[] Question: Does the paper describe potential risks incurred by study participants, whether such risks were disclosed to the subjects, and whether Institutional Review Board (IRB) approvals (or an equivalent approval/review based on the requirements of your country or institution) were obtained?
    \item[] Answer: \answerNA{}.
    \item[] Justification: The paper does not involve research with human subjects, so IRB approval or equivalent review is not applicable.

\item {\bf Declaration of LLM usage}
    \item[] Question: Does the paper describe the usage of LLMs if it is an important, original, or non-standard component of the core methods in this research? Note that if the LLM is used only for writing, editing, or formatting purposes and does \emph{not} impact the core methodology, scientific rigor, or originality of the research, declaration is not required.
    \item[] Answer: \answerYes{}.
    \item[] Justification: LLMs were used for coding and debugging assistance, but not for generating data, running evaluations, selecting results, or forming any core methodological component of the research.

\end{enumerate}
\appendix

\section{Additional Training and Data Generation Details}
\label{app:training}

\paragraph{Grammar generation.}
Alphabet size is $N=20$. Each character has between 1 and 3 right-hand sides drawn uniformly.
Right-hand-side lengths follow a truncated Zipfian distribution over $[2,6]$.
All right-hand sides are globally unique.

\paragraph{Chain generation.}
Starting-word lengths are sampled uniformly from $[2,12]$ and chain lengths from $[5,50]$.
For the default setup, the contraction weight is $\rho = 4$.

\paragraph{Model and optimization.}
The pretrained model uses 12 layers, hidden dimension 512, 8 attention heads, an MLP expansion ratio
of 4, RoPE positional encoding, QK normalization, and a $\tanh$ soft-cap on logits.
Optimization uses AdamW with a warm/flat/cool schedule, peak learning rate $8\times10^{-4}$,
sequence length 512, bfloat16 training, and gradient clipping at norm 1.0.

\paragraph{Pretraining data generation.}
For each pretraining chain, we first sample a starting word length and a chain
length from the configured distributions. Starting from a random word over the
alphabet, the generator repeatedly enumerates all currently valid primitive
moves: expansions at characters whose source symbol appears in the grammar, and
contractions at substrings matching a right-hand side. One move is sampled from
the configured rule weights, applied to the current word, and serialized as
\[
\texttt{lhs\_word | action | rhs\_word}.
\]
The resulting \texttt{rhs\_word} becomes the next \texttt{lhs\_word}. Thus
pretraining chains are stochastic local rewrite trajectories, not
solution-directed demonstrations. The generator records the start-word
distribution, chain-length distribution, rule-weight distribution, contraction
weight, and matched-fraction setting used for each dataset.

\paragraph{Post-training prompt generation.}
For RL/RFT post-training, prompts are generated by sampling a target symbol
$c^\star$ and applying forward expansions to obtain a prompt word
$w_{\mathrm{prompt}}$. The model sees only $w_{\mathrm{prompt}}$ and the target;
the expansion path used to construct the prompt is not provided. Prompts are
filtered by the primitive contraction budget: Difficulty~1 fits within the
generation budget, while higher difficulties require additional primitive
contraction steps beyond the budget. The same prompt distribution is used for
GRPO and RFT; the methods differ only in how sampled completions are used for
optimization.

Unless otherwise stated, learning curves are averaged over three training seeds

\paragraph{Compute.}
All experiments were run on a node with $4\times$ NVIDIA H100 GPUs.

\section{GRPO and RFT Details}
\label{app:grpo}

\paragraph{Reward.} The reward is
\[
r =
\begin{cases}
1 & \text{if the trajectory is well-formed and ends with } c^\star, \\
0 & \text{otherwise.}
\end{cases}
\]
A trajectory is \emph{well-formed} if it is parseable as a sequence of
\texttt{lhs\_word | action | rhs\_word} triples and each triple's
\texttt{rhs\_word} matches the next triple's \texttt{lhs\_word}. This check is
syntactic and does not require each intermediate rewrite to be grammar-valid.

\paragraph{GRPO.}
For each prompt $x_i$, GRPO samples $G$ completions $\{y_{i,g}\}_{g=1}^G$
and assigns each a scalar reward $r_{i,g}$. Advantages are normalized within
each prompt group:
\[
A_{i,g} = \frac{r_{i,g} - \mu_i}{\sigma_i + \delta},
\qquad
\mu_i = \frac{1}{G}\sum_{g=1}^{G} r_{i,g},
\qquad
\sigma_i = \sqrt{\frac{1}{G}\sum_{g=1}^{G}(r_{i,g}-\mu_i)^2}.
\]
Letting
\[
\rho_{i,g,t}(\theta)
=
\frac{\pi_\theta(y_{i,g,t}\mid x_i, y_{i,g,<t})}
{\pi_{\mathrm{old}}(y_{i,g,t}\mid x_i, y_{i,g,<t})},
\]
the objective is
\[
\mathcal{L}(\theta)
=
-\frac{1}{\sum_{i,g,t} m_{i,g,t}}
\sum_{i,g,t} m_{i,g,t}
\left[
\min\!\left(
\rho_{i,g,t} A_{i,g},
\mathrm{clip}(\rho_{i,g,t}, 1-\epsilon, 1+\epsilon) A_{i,g}
\right)
-\beta \kappa_{i,g,t}(\theta)
\right],
\]
where $m_{i,g,t}$ masks prompt and post-stop tokens and $\kappa_{i,g,t}$ is
a reverse-KL penalty to the frozen pretrained reference policy. In our
experiments, $G=4$, $\epsilon=0.2$, $\beta=10^{-3}$, temperature is $0.8$,
top-$k$ is $5$, and the maximum generation length is $256$ tokens.

\paragraph{RFT.}
RFT uses the same prompt distribution, group size $G=4$, batch size,
optimizer, and KL anchor as GRPO. For each prompt, it samples $G$
completions, retains those with $r=1$, and performs next-token prediction
on the accepted rollouts using the same loss masking $m_{i,g,t}$. RFT
differs from GRPO only in (i) it discards rather than down-weights failed
rollouts, and (ii) its update is supervised rather than policy-gradient.

\section{Finite-Sample Selection in RFT and GRPO}
\label{app:feature_selection}

This appendix formalizes the feature-level selection argument used in
Section~\ref{sec:structured_exploration}. The goal is not to model the full
Transformer update exactly, but to isolate a finite-sample distinction between
GRPO and RFT. GRPO compares successful and failed completions sampled for the
same prompt. RFT trains only on accepted completions.

\paragraph{Finite-group contrast in GRPO.}
Consider a binary trajectory feature $F(\tau)\in\{0,1\}$ indicating that a
sampled trajectory contains a particular action pattern, such as a macro
contraction, a parallel contraction, or a spurious contraction. For one prompt,
GRPO samples a group of $G$ completions. Let $n$ be the number of successful
completions, so $G-n$ completions fail. Let $k^+$ be the number of successful
completions containing the feature, and $k^-$ the number of failed completions
containing the feature.

For binary rewards, successful completions have reward $1$ and failed
completions have reward $0$. Within this prompt group, the mean reward is
\[
\mu = \frac{n}{G}.
\]
The reward standard deviation is
\[
\sigma
=
\sqrt{
\frac{1}{G}
\left[
n\left(1-\frac{n}{G}\right)^2
+
(G-n)\left(0-\frac{n}{G}\right)^2
\right]
}
=
\frac{\sqrt{n(G-n)}}{G}.
\]
Therefore, for groups with $0<n<G$, the normalized advantage of a successful
completion is
\[
A^+
=
\frac{1-\mu}{\sigma}
=
\frac{1-\frac{n}{G}}{\frac{\sqrt{n(G-n)}}{G}}
=
\sqrt{\frac{G-n}{n}},
\]
and the normalized advantage of a failed completion is
\[
A^-
=
\frac{0-\mu}{\sigma}
=
-\sqrt{\frac{n}{G-n}}.
\]

Now sum the advantage contribution over trajectories containing feature $F$.
Ignoring token-level constants, clipping effects, and the KL term, the
feature-count contribution is proportional to
\[
A^+k^+ + A^-k^-.
\]
Substituting the expressions above gives
\[
A^+k^+ + A^-k^-
=
\sqrt{\frac{G-n}{n}}\,k^+
-
\sqrt{\frac{n}{G-n}}\,k^-
=
\sqrt{n(G-n)}
\left(
\frac{k^+}{n}
-
\frac{k^-}{G-n}
\right).
\]
The prefactor $\sqrt{n(G-n)}$ is positive and therefore does not affect the
sign. Thus the sign of the feature-level GRPO pressure is determined by
\[
\frac{k^+}{n}
-
\frac{k^-}{G-n}.
\]
GRPO increases features that are more frequent among successful completions
than failed completions from the same prompt, and suppresses features enriched
among failed completions. Groups with all successes or all failures have no
binary reward contrast under this normalization.

\paragraph{Why this matters for spurious actions.}
Suppose a spurious contraction appears in some successful trajectories, but also
appears frequently in failed trajectories for the same prompt. RFT sees only
the successful trajectories and treats the spurious action as part of the target
sequence. GRPO instead compares success-side and failure-side frequencies. If
the spurious action is not more common in successes than failures, it receives
little positive pressure or even negative pressure. This is the finite-sample
``passenger action'' distinction: RFT clones accepted traces wholesale, while
GRPO has a same-prompt contrast that can suppress patterns that travel with
failure.

\paragraph{RFT lacks the same-group failure-side term.}
For the same finite group, RFT keeps only successful completions. The empirical
feature frequency available to RFT is therefore
\[
\hat f^+ = \frac{k^+}{n}.
\]
The failed-completion term $k^-/(G-n)$ does not appear in the supervised
targets. As a result, a spurious action inside an accepted trajectory is fit
alongside useful actions, even if the same action pattern is equally common or
more common in failed rollouts.

\paragraph{Why RFT is still a strong baseline.}
This finite-sample distinction should not be read as saying that RFT is
intrinsically misaligned with reward optimization. In the infinite-sample,
infinitesimal-update limit, on-policy RFT with binary rewards is first-order
aligned with the policy gradient.

Let
\[
J(\theta)=\mathbb{E}_{\tau\sim\pi_\theta}[R(\tau)].
\]
Using the score-function identity,
\[
\nabla_\theta J(\theta)
=
\mathbb{E}_{\tau\sim\pi_\theta}
\left[
R(\tau)\nabla_\theta \log \pi_\theta(\tau)
\right].
\]
For binary rewards, this becomes
\[
\nabla_\theta J(\theta)
=
P_\theta(R=1)
\,
\mathbb{E}_{\tau\sim\pi_\theta(\tau\mid R=1)}
\left[
\nabla_\theta \log \pi_\theta(\tau)
\right].
\]
On-policy RFT samples from $\pi_{\theta_{\mathrm{old}}}$, keeps successful
trajectories, and minimizes
\[
\mathcal{L}_{\mathrm{RFT}}(\theta)
=
-\mathbb{E}_{\tau\sim\pi_{\theta_{\mathrm{old}}}(\tau\mid R=1)}
[\log\pi_\theta(\tau)].
\]
Evaluated at $\theta=\theta_{\mathrm{old}}$,
\[
-\nabla_\theta \mathcal{L}_{\mathrm{RFT}}(\theta_{\mathrm{old}})
=
\mathbb{E}_{\tau\sim\pi_{\theta_{\mathrm{old}}}(\tau\mid R=1)}
\left[
\nabla_\theta \log \pi_\theta(\tau)
\right]_{\theta=\theta_{\mathrm{old}}}
=
\frac{1}{P(R=1)}
\nabla_\theta J(\theta_{\mathrm{old}}).
\]
Thus ideal on-policy RFT points in the same first-order direction as
binary-reward policy gradient, up to a positive scalar. This explains why RFT
is a strong early baseline in our experiments. The later divergence arises in
the finite iterative setting: RFT repeatedly clones finite accepted
trajectories, while GRPO uses negative advantages from below-average
completions in the same prompt group.

\paragraph{Connection to score-function selection.}
The finite-group expression above is a sample-level version of the standard
score-function selection signal. For any trajectory feature $\phi(\tau)$ and
binary reward $R(\tau)$,
\[
\operatorname{Cov}_{\pi}(\phi,R)
=
P_{\pi}(R=1)P_{\pi}(R=0)
\left(
\mathbb{E}_{\pi}[\phi\mid R=1]
-
\mathbb{E}_{\pi}[\phi\mid R=0]
\right).
\]
Thus features enriched in successful trajectories receive positive
reward-gradient pressure, while features enriched in failed trajectories receive
negative pressure. The main text uses the finite-group version because it is
the contrast directly implemented by GRPO.

\section{Pretraining Contraction Weighting}
\label{app:rho_pretraining}

The contraction weight $\rho$ is a local weight multiplier, not a direct
contraction-frequency parameter. At each pretraining step, the generator
enumerates the valid primitive expansion moves $E(w)$ and primitive contraction
moves $C(w)$ from the current word $w$. Expansion moves keep their base weights,
while contraction moves are multiplied by $\rho$ before sampling the next
rewrite.

For example, suppose the current word admits four valid primitive moves: three
expansions and one contraction, all with unit base weight. With $\rho=1$, all
four moves have equal weight, so the contraction probability is
\[
\frac{1}{3+1}.
\]
With $\rho=4$, the three expansions still have weight $1$ each, but the
contraction has weight $4$, so the contraction probability becomes
\[
\frac{4}{3+4}.
\]
Thus increasing $\rho$ makes contractions more likely locally whenever
contractions are available. It does not directly impose a fixed global
contraction rate.

More generally, if each move $a$ has base rule weight $q(a)$, then the
probability that the next move is a contraction is
\[
P_\rho(C \mid w)
=
\frac{
\rho\sum_{a\in C(w)} q(a)
}{
\sum_{a\in E(w)}q(a)
+
\rho\sum_{a\in C(w)}q(a)
}.
\]
The realized fraction of contractions over the full pretraining corpus depends
on the grammar, the current word distribution, which moves are available, and
the trajectory dynamics. This is why $\rho$ should be interpreted as a local
contraction weight rather than as the realized contraction rate.

High $\rho$ also changes how contractions are arranged. Since contraction moves
are locally upweighted whenever a reducible state is reached, the generator is
more likely to remain in a reduction mode and produce consecutive contraction
steps. Thus high $\rho$ exposes the model not only to more contractions in
aggregate, but also to more chained primitive reductions.

\paragraph{Matched-fraction control.}
The matched-fraction control separates aggregate contraction exposure from this
local weighting mechanism. We choose a low-$\rho$ generator and adjust its
direction-sampling probability so that its realized marginal contraction rate is
comparable to the $\rho=2$ pretraining run. After choosing whether the next move
is an expansion or contraction, a valid move within that direction is sampled.

The control is therefore not an unbiased $\rho=1$ generator. It is intentionally
biased to match the marginal number of contraction examples in an intermediate
contraction-weight setting. What it removes is the local reweighting of every
available contraction move by a larger $\rho$. Consequently, contractions occur
at a comparable overall rate, but they are less likely to appear as sustained
contraction chains.

This distinction is what Figure~\ref{fig:rho} tests. If strategy emergence were
explained only by the marginal number of contraction examples, the matched
control should behave like the $\rho=2$ model. Instead, it behaves much closer
to the low-$\rho$ model, suggesting that the organization of contractions across
trajectories---especially exposure to reduction-mode behavior and chained
primitive contractions---matters for later strategy emergence under RL.

\section{Example Grammar and Pretraining Chains}

We used the following grammar instance for the data generation. Each source
symbol maps to one or more globally unique right-hand-side strings:
\[
\begin{aligned}
a &\rightarrow \{\texttt{qs}\}, &
b &\rightarrow \{\texttt{ic}, \texttt{fe}, \texttt{sqjg}\}, &
c &\rightarrow \{\texttt{jaocn}\}, \\
d &\rightarrow \{\texttt{ba}, \texttt{nklt}, \texttt{si}\}, &
e &\rightarrow \{\texttt{mj}\}, &
f &\rightarrow \{\texttt{hl}, \texttt{mq}\}, \\
g &\rightarrow \{\texttt{hheoj}, \texttt{jd}, \texttt{mi}\}, &
h &\rightarrow \{\texttt{aps}, \texttt{gna}\}, &
i &\rightarrow \{\texttt{paolh}\}, \\
j &\rightarrow \{\texttt{qsn}, \texttt{cnr}, \texttt{gd}\}, &
k &\rightarrow \{\texttt{csj}\}, &
l &\rightarrow \{\texttt{dff}\}, \\
m &\rightarrow \{\texttt{fc}, \texttt{pt}\}, &
n &\rightarrow \{\texttt{dl}, \texttt{hfd}, \texttt{nq}\}, &
o &\rightarrow \{\texttt{cj}, \texttt{hq}\}, \\
p &\rightarrow \{\texttt{fk}\}, &
q &\rightarrow \{\texttt{imr}, \texttt{nd}\}, &
r &\rightarrow \{\texttt{gn}\}, \\
s &\rightarrow \{\texttt{iq}, \texttt{bb}, \texttt{fflgj}\}, &
t &\rightarrow \{\texttt{ld}, \texttt{ka}\}. &&
\end{aligned}
\]

A pretraining example is a stochastic chain of local primitive rewrites. Each
line applies one valid expansion or contraction, and the output word becomes the
input to the next line. For example:
\begin{center}
\begin{minipage}{0.62\linewidth}
\small
\begin{verbatim}
bqt | 0 E b>fe | feqt
feqt | 2 E q>nd | fendt
fendt | 4 E t>ka | fendka
fendka | 0 1 C b>fe | bndka
bndka | 0 E b>sqjg | sqjgndka
sqjgndka | 6 E k>csj | sqjgndcsja
\end{verbatim}
\end{minipage}
\end{center}

Another sampled snippet is:
\begin{center}
\begin{minipage}{0.62\linewidth}
\small
\begin{verbatim}
ltm | 0 E l>dff | dfftm
dfftm | 3 E t>ld | dffldm
dffldm | 5 E m>pt | dffldpt
dffldpt | 3 4 C t>ld | dfftpt
dfftpt | 3 E t>ka | dffkapt
dffkapt | 5 E p>fk | dffkafkt
\end{verbatim}
\end{minipage}
\end{center}

These snippets illustrate the local nature of pretraining: the model sees
grammar-consistent primitive rewrites, not the macro or parallel contractions
analyzed in the main text.

\subsection{Example RL/RFT Post-Training Prompts}

The same prompt distribution is used for RL and RFT post-training. For RL, the
model samples completions for each prompt and receives a binary reward. For
RFT, successful on-policy completions from the same prompts are retained as
next-token prediction targets. The model sees only the final expanded prompt
word and the target symbol; the expansion chains below are shown only for
exposition.

For readability, the examples below use an illustrative primitive-step budget
$B=3$ rather than the 256-token serialized budget used in the experiments.
Thus Difficulty~1 fits within the budget, Difficulty~2 requires one additional
primitive contraction step, and Difficulty~3 requires two additional primitive
contraction steps.

\paragraph{Difficulty 1: primitive solution fits within budget.}
Starting from target symbol \texttt{a}, one possible prompt construction is:
\begin{center}
\begin{minipage}{0.5\linewidth}
\small
\begin{verbatim}
a | 0 E a>qs | qs
qs | 0 E q>nd | nds
nds | 2 E s>iq | ndiq
\end{verbatim}
\end{minipage}
\end{center}

The post-training prompt is therefore \texttt{ndiq}, with target \texttt{a}.
A primitive contraction solution is:
\begin{center}
\begin{minipage}{0.5\linewidth}
\small
\begin{verbatim}
ndiq | 0 1 C q>nd | qiq
qiq | 1 2 C s>iq | qs
qs | 0 1 C a>qs | a
\end{verbatim}
\end{minipage}
\end{center}

This solution uses three primitive contractions, so it fits within the
illustrative budget $B=3$.

\paragraph{Difficulty 2: one primitive step beyond budget.}
Starting from target symbol \texttt{b}, one possible prompt construction is:
\begin{center}
\begin{minipage}{0.62\linewidth}
\small
\begin{verbatim}
b | 0 E b>sqjg | sqjg
sqjg | 0 E s>bb | bbqjg
bbqjg | 2 E q>imr | bbimrjg
bbimrjg | 5 E j>gd | bbimrgdg
\end{verbatim}
\end{minipage}
\end{center}

The post-training prompt is \texttt{bbimrgdg}, with target \texttt{b}.
A primitive contraction solution is:
\begin{center}
\begin{minipage}{0.62\linewidth}
\small
\begin{verbatim}
bbimrgdg | 2 4 C q>imr | bbqgdg
bbqgdg | 3 4 C j>gd | bbqjg
bbqjg | 0 1 C s>bb | sqjg
sqjg | 0 3 C b>sqjg | b
\end{verbatim}
\end{minipage}
\end{center}

This requires four primitive contractions, one step beyond the illustrative
budget.

\paragraph{Difficulty 3: two primitive steps beyond budget.}
Starting from target symbol \texttt{t}, one possible prompt construction is:
\begin{center}
\begin{minipage}{0.72\linewidth}
\small
\begin{verbatim}
t | 0 E t>ka | ka
ka | 0 E k>csj | csja
csja | 0 E c>jaocn | jaocnsja
jaocnsja | 5 E s>iq | jaocniqja
jaocniqja | 7 E j>gd | jaocniqgda
\end{verbatim}
\end{minipage}
\end{center}

The post-training prompt is \texttt{jaocniqgda}, with target \texttt{t}.
A primitive contraction solution is:
\begin{center}
\begin{minipage}{0.72\linewidth}
\small
\begin{verbatim}
jaocniqgda | 0 4 C c>jaocn | ciqgda
ciqgda | 1 2 C s>iq | csgda
csgda | 2 3 C j>gd | csja
csja | 0 2 C k>csj | ka
ka | 0 1 C t>ka | t
\end{verbatim}
\end{minipage}
\end{center}

This requires five primitive contractions, two steps beyond the illustrative
budget.

These examples illustrate the post-training data construction. RL and RFT use
the same prompts and final-answer reward; they differ only in how sampled
rollouts are used for optimization. GRPO updates from both successful and
failed completions through group-relative advantages, while RFT retains only
successful completions as supervised targets.

\section{Boolean Circuit Experiment Details}
\label{app:boolean}

\paragraph{Task.}
A problem is a Boolean circuit written as a sequence of gate assignments, each
computing a new wire from input variables or earlier wires. The model must
reduce the circuit to its simplest form by emitting a step-by-step
simplification trace. A primitive step eliminates a single gate: given
$g_1 = B \oplus B$ and $g_2 = A \oplus g_1$, a primitive trace first derives
$g_1 = 0$ and then $g_2 = A$. A macro step collapses several valid eliminations
into one, here $A \oplus (B \oplus B) \rightarrow A$. Gate elimination thus
plays the role of primitive contraction in the grammar, and multi-gate collapse
the role of macro contraction.

\paragraph{Controlling what the model is taught.}
Gate types are presented to the model as opaque tokens rather than names such as
XOR or AND, so it cannot rely on their pretrained meaning; readable names are
used in the paper only for exposition. Variable and wire names are resampled for
every problem, so no macro recurs as the same token string, and evaluation
circuits are disjoint from training circuits. An initial SFT stage teaches the
hidden gate semantics, the trace format, and single-gate eliminations. None of
the macro schemas analyzed in Section~\ref{sec:boolean} is demonstrated during
SFT.

\paragraph{Post-training.}
Starting from the SFT checkpoint, we run GRPO with only a final-answer reward and
a generation budget of 128 tokens.

\paragraph{Difficulty and evaluation.}
As in the grammar, difficulty is defined relative to the generation budget. D0
circuits admit a primitive solution within 128 tokens, while D1--D4 circuits
require compressing one, two, three, and four primitive gate lines,
respectively, to fit within the budget. We evaluate on 40 held-out circuits per
bucket (200 in total) with identical sampling settings for both policies, and
report pass@1024 for the SFT policy and pass@16 for GRPO.

\begin{table}[!ht]
\centering
\small
\caption{Macro schemas in GRPO solutions on held-out circuits. $X$ and $Y$
stand for any subexpression.}
\label{tab:boolean_schemas}
\begin{tabular}{lcccc}
\toprule
Schema & In SFT & Required in RL & Correct & Bindings \\
\midrule
$X \oplus (Y \oplus Y) \rightarrow X$      & No & Yes & 20/20 & 20 \\
$X \land (Y \oplus \lnot Y) \rightarrow X$ & No & Yes & 22/22 & 22 \\
$X \lor (Y \oplus Y) \rightarrow X$        & No & No  & 7/9   & 7  \\
$(X \land Y) \lor X \rightarrow X$         & No & No  & 0/11  & 0  \\
\bottomrule
\end{tabular}
\end{table}

\paragraph{Auditing.}
Every intermediate assignment in a generated trace is checked independently with
exact Boolean algebra, and a solution counts only if all steps are valid and the
final form is correct. For GRPO, the number of audited successes equals the
number of final-answer-only successes in every bucket, so no success relies on
an invalid step.

\paragraph{Schema analysis.}
We group valid macro steps into schemas by normalizing their variable names, so
that applications of the same operation under different variable bindings map
to the same schema. Table~\ref{tab:boolean_schemas} reports, for each schema,
the evaluation circuits on which GRPO applies it correctly out of those where it
applies, and the number of distinct variable bindings among the correct
applications. The first three schemas all collapse a subexpression that
evaluates to a constant and then eliminate it. Absorption,
$(X \land Y) \lor X \rightarrow X$, is a structurally different simplification
and is never applied, which bounds the transfer reported in the main text.

\end{document}